\documentclass[letterpaper, 10 pt, conference]{ieeeconf}  % Comment this line out if you need a4paper

\IEEEoverridecommandlockouts                              % This command is only needed if 
\usepackage{graphics} % for pdf, bitmapped graphics files
\usepackage{epsfig} % for postscript graphics files
\usepackage{mathptmx} % assumes new font selection scheme installed
\usepackage{times} % assumes new font selection scheme installed
\usepackage{amsmath} % assumes amsmath package installed
\usepackage{amssymb}  % assumes amsmath package installed
\usepackage{graphicx}
\usepackage{float}
\usepackage{tabularx}
\usepackage{arydshln}
\usepackage{subcaption}
\usepackage{dirtytalk}
\usepackage[
    natbib=true,
    style=numeric,
    sorting=none
]{biblatex}

\title{\LARGE \bf
Robot Vitals and Robot Health: Towards Systematically Quantifying Runtime Performance Degradation in Robots Under Adverse Conditions
}

\author{Aniketh Ramesh$^{1}$, Rustam Stolkin$^{1}$,  and Manolis Chiou$^{1}$% <-this % stops a space
\thanks{This work was supported by NCNR EP/R02572X/1, EP/P01366X/1, EP/P017487/1, and ReLiB FIRG005.}% <-this % stops a space
\thanks{$^{1}$Extreme Robotics Lab (ERL) and National Center for Nuclear Robotics (NCNR), University of Birmingham, UK}
\thanks{\tt axr1050@student.bham.ac.uk\tt, m.chiou@bham.ac.uk}%
}

\newcolumntype{b}{X}
\newcolumntype{s}{>{\hsize=.5\hsize}X}

\begin{document}

\maketitle
\thispagestyle{empty}
\pagestyle{empty}

%%%%%%%%%%%%%%%%%%%%%%%%%%%%%%%%%%%%%%%%%%%%%%%%%%%%%%%%%%%%%%%%%%%%%%%%%%%%%%%%
\begin{abstract}

This paper addresses the problem of automatically detecting and quantifying performance degradation in remote mobile robots during task execution. A robot may encounter a variety of uncertainties and adversities during task execution, which can impair its ability to carry out tasks effectively and cause its performance to degrade. Such situations can be mitigated or averted by timely detection and intervention (e.g., by a remote human supervisor taking over control in teleoperation mode). Inspired by patient triaging systems in hospitals, we introduce the framework of \say{robot vitals} for estimating overall \say{robot health}. A robot's vitals are a set of indicators that estimate the extent of performance degradation faced by a robot at a given point in time. Robot health is a metric that combines robot vitals into a single scalar value estimate of performance degradation. Experiments, both in simulation and on a real mobile robot, demonstrate that the proposed robot vitals and robot health can be used effectively to estimate robot performance degradation during runtime.
\end{abstract}

%%%%%%%%%%%%%%%%%%%%%%%%%%%%%%%%%%%%%%%%%%%%%%%%%%%%%%%%%%%%%%%%%%%%%%%%%%%%%%%%
\section{Introduction}

Robots operating in remote environments, regardless of their advanced capabilities, often face various issues and performance degradation during task execution. This is particularly true for tasks in extreme environments, where the robots are typically physically remote from a human operator who remains in a safe zone. Examples of performance degrading factors include terrain adversities, camera occlusion, sensor noise, limitations in AI capabilities, and unexpected circumstances. Irrespective of the control mode used (e.g., full autonomous/ Teleoperated Robots, Shared Control, Variable autonomy), when robots are subjected to such factors for a prolonged period, they may behave unpredictably, perform tasks sub-optimally, or fail catastrophically.

Robot performance degradation is commonly mitigated by changing the control mode, or by initiating pre-programmed recovery behaviours. While it is possible for a remote human supervisor to detect problems and intervene \cite {Chiou2016IROS_HI}, it would also be useful to provide robots with a means of reliable and automatic detection of performance degradation. This would enable autonomous robots to trigger automatic recovery behaviours or call for human intervention \cite{chiou2019mixed}.

In this study, the term \say{performance degradation} refers to any impairment in the capability of a robot to carry out its tasks. Automatically detecting situations where a robot is facing performance degradation in real time, is a challenging open problem. Any such framework requires metrics applicable to a wide range of robots; to quantify the effect of hardware errors, software limitations, and environmental factors etc. during problematic situations. The framework should also be robust, and easily adaptable to constraints imposed by the robot's sensors, morphology or underlying algorithms used. 

This paper proposes a framework to detect and quantify robot performance degradation during task execution, simultaneously for different levels of abstraction, by using a set of performance indicators called the \say{robot vitals}. The \say{robot health} is a meta-metric that combines all vitals into a single scalar value, estimating the intensity of performance degradation. Using such a framework, an AI agent can estimate the performance degradation that a robot is facing (e.g., via visual cues \cite{humphrey2007assessing}), call for operator assistance for semi-autonomous robots \cite{murphy2010navigational}, trigger recovery behaviours automatically, or even suggest actions to mitigate the effect of problematic situations. The design of this framework is inspired by the simplicity and standardisation of patient triaging systems used by hospitals, where threshold values of the classic \say{vital signs} physiological parameters (e.g., pulse, temperature, blood pressure, and oxygen saturation) to assign a clinical health intensity score to patients \cite{smith2019national}.

Estimating performance degradation can also inform AI policies on shared control and/or mixed-initiative control systems (i.e., humans and robots capable of seizing or relinquishing control of different mission elements \cite{Jiang2015, chiou2019mixed}). This is particularly important in multi-robot systems, as the operator is required to simultaneously monitor several robots with their limited cognitive resources, leading to high cognitive workload and suboptimal assistance \cite{kolling2015human}. For these reasons, our experiments test the framework on a mobile robot that is navigating fully autonomously throughout each experiment, while encountering challenging situations. However, our intention is that this system can then be used (in future work) to trigger autonomous recovery behaviours or inform Level of Autonomy (LoA) switching. We hypothesise that by using robot vitals and robot health, the runtime performance degradation of robots can be estimated.

The main contributions of this paper are a) proposing and introducing the framework of robot vitals and robot health; b) proposing a realisation with a set of 5 robot vitals and an intuitive scalable meta-metric that combines the robot vitals to calculate a robot's health; c) presenting and examining the results of systematic experiments on the robot vitals and robot health framework.

\section{Related Work} \label{sec:RelatedWork}

Robot performance degradation that can be mitigated without removing a robot from the operating environment, is called a field repairable failure \cite{carlson2005ugvs}. Such failures are non-terminal \cite{carlson2005ugvs} as they only cause a temporary lapse in task execution. However, if left unattended for long, non-terminal failures can become terminal. Some factors responsible for such failures are sensor noise, unpredictable robot behaviour, wheel encoder faults, motor malfunctions, communication losses and sudden power drops \cite{tsarouhas2016mission,honig2018understanding, steinbauer2012survey}.

Literature on performance evaluation commonly describes metrics that can be calculated offline before or after the robot has finished operating. These metrics quantify a robot's task performance (e.g., time to complete a mission, area coverage, or victims identified), a robot's reliability (e.g., Mean Time to Failure and Mean Time Between Failures) \cite{khalastchi2018fault, honig2018understanding}, or the qualitative risk (low, medium, or high risk) posed by a robot to its operators and surroundings \cite{brooks2017human,garza2018failure}. In contrast, our work offers the advantage of online detection and quantification of robot performance during problematic situations.

The most common method of online detection of hardware or electronic issues related to robot movement are control systems and dead-reckoning systems \cite{yu2011particle}. Such systems trigger an error signal if a robot deviates from a pre-defined model of ideal robot behaviour. Alternatively, runtime performance degradation can be detected using heuristics defined on threshold values of sensor data. These heuristics may use the difference between localisation estimates \cite{mendoza2012mobile}, the difference between the robot’s actual velocity and the ideal velocity given by an expert planner \cite{chiou2019mixed}, the difference between the expected time to complete a task \cite{mendoza2012mobile,wegner2006agent} and the actual time taken, a robot's mean velocity, displacement or the total area explored \cite{valero2008adaptative}. Alternatively, online metrics can also be learnt through machine learning or reinforcement learning \cite{doroodgar2014learning, hong2019investigating} but require large data sets of robot failures \cite{dua2017uci} to be available or created in simulation. In summary, the papers cited above use task-specific online metrics to detect poor robot performance. They do not propose a general framework for measuring degradation and do not explicitly examine the relation between the metrics and robot performance degradation.

Our previous work \cite{ramesh2021robot} gave a rudimentary introduction to the concept of robot vitals and robot health. Here we present a systematic approach to reason about the vitals, and an experimental validation of the health metric. To the best of our knowledge, there is a gap in the literature on experimentally validated approaches to quantifying robot performance degradation online in real-time. We aim to address this gap by proposing a framework to quantify runtime performance degradation using a set of performance indicators and a meta-metric, while validating the framework with systematic experiments. 

\section{Robot Vitals}\label{sec:RV}

Robot vitals are a set of metrics that indicate the performance degradation faced by a robot at any given time. In contrast to task performance metrics, vitals indicate a robot's ability to function during adverse conditions without failing or behaving erroneously. Each vital represents a specific aspect of robot behaviour during adverse conditions. While no single vital may give definitive information that a robot is failing, observing the trend of a set of vitals may serve as a robust indicator of adverse conditions and may help understand the nature of adversities that robots encounter. Ideally, a set of robot vitals should account for all the performance degrading factors encountered in the environment of operation. 

A robot is \say{suffering} if it is experiencing high performance degradation, making it likely to perform sub-optimally or break down. Different aspects of performance degradation are related to their corresponding vitals using the probability of robot \say{suffering} given each vital. Hence, as the performance degradation indicated by a vital increases, the probability of suffering given the vital should increase. Each vital is calculated by filtering and sampling real-time data from robot operation. Next, transforms like event detection or thresholding are applied on the vitals to emphasise features of interest. Finally, the probability of suffering for each vital is calculated as a function of the resultant sequence of values. 

As one possible and indicative realisation, we present a set of five vitals which are useful in our experiments with remotely operating variable autonomy mobile robots, and the rational to derive them. The first four vitals capture the motion-related performance degradation. The fifth captures localisation-related performance degradation. The probability distribution function and transforms used for each vital are determined empirically based on preliminary experiments and observations of multiple mobile robotic platforms, and previous work \cite{chiou2015towards,Chiou2016IROS_HI,chiou2019mixed}. These vitals are not meant to be an exhaustive list. Depending on the robotic platform, task, and the environment, a variety of metrics can be chosen as robot vitals by adhering to above mentioned principles.

\subsection{Rate of Change of Distance from navigational goal ($\dot{d}_{g}$)}

This vital is used to indicate situations in which performance degrading factors cause a robot to not move towards its navigational goal. Such situations can be detected by observing the Rate of Change (RoC) of distance from a robot's current position to its current navigational goal ($\dot{d}_{g}$). The odometry position estimate obtained after Extended Kalman Filter sensor fusion is used as the robot's current position, and the goal is given by an operator or the navigation algorithm. The ${d}_{g}$ is calculated using Euclidean distance to make minimum assumptions about the task, the algorithm used, and whether the map is known before the task. However, for more sophisticated applications ${d}_{g}$ can be calculated using the distance remaining along a non-linear path. During little to no performance degradation (i.e., ideal behaviour), the robot moves towards the goal with uniform velocity. This results in a constant value $\dot{d}_{g}<0$, barring few fluctuations. A $\dot{d}_{g} \approx 0$ has very little similarity to ideal behaviour and indicates that a robot is unable to move. Lastly, $\dot{d}_{g} > 0$ is dissimilar to ideal behaviour, and indicates that performance degradation has resulted in the robot taking a sub-optimal path or moving away from the goal. 
To calculate the magnitude of similarity (${d}_{event}$), we observe $\dot{d}_{g}$ values over multiple time steps, and compare it to ideal behaviour using a convolutional matched filter \cite{turin1995introduction}. Preliminary experiments showed that ${d}_{event}>0.3$ indicates a high degree of similarity i.e., the robot is facing little to no performance degradation. A ${d}_{event}<-0.3$ indicates dissimilarity and suggests that the robot is unable to move or is moving away from the goal. The probability of suffering is calculated as a function of ${d}_{event}$ such that its value is high if  ${d}_{event}<-0.3$ and low if ${d}_{event}>0.3$. To increase the sensitivity of the probability distribution to $d_{event} \in [-1,1]$,  we use the sigmoid function given below with constants $a=-6$ and $b=-0.15$:

\begin{align}
    P(\textit{suffering}|\dot{d}_{g}, \dot{d}_{event}) = \frac{1}{1+\exp\left(\left(-a \cdot d_{event}+a\cdot b\right)\right)}
    \label{eq: sigmoid_d}
\end{align}

\subsection{Jerk along Axis of Motion ($\dot{a}_z$)}

This vital detects situations in which performance degrading factors like uneven terrain may result in sudden jerks or jittering along the axis of motion (z axis generally). Sudden dips in terrain elevation can rapidly increase the force on one side, thereby causing the robot to tilt or topple. A higher magnitude of jerk indicates that the robot is more likely to topple, making the probability of suffering higher. During preliminary experiments with a simulated Clearpath Husky robot, we observed that sudden jerks of $\pm$ 30 degrees ($\approx \pm 0.5$ radians) or above along the z-axis may increase the likelihood of the robot toppling over. Therefore, the probability of suffering given jerk along the Z-axis should be high when $\mid \dot{a}_z \mid \approx \pm 0.5$ radians, and low if $\mid \dot{a}_z \mid \approx \pm 0$. 

The jerk magnitude along the axis of motion is calculated using the rate of change of linear acceleration along the Z axis $\dot{a}_z$. $a_z$ is usually measured using an Inertial measurement unit (IMU). Since IMU readings tend to be noisy, raw IMU output values smoothened using a rolling window average and then downsampled to one reading per second before calculating $\dot{a}_z$. The function for $P(\textit{suffering}| \dot{a}_{z})$ is calculated using an inverted bell curve as follows:

\begin{align}
    P(\textit{suffering}| \dot{a}_{z}) = 1\ -\ \frac{1}{\left(2\pi\right)^{\frac{1}{2}}\sigma_{1}} e^{\left(-\left(\frac{0.5}{\sigma_{2}^{2}}\left(\dot{a}_{z} \right)^{2}\right)\right)}
\end{align}

The values of $\sigma1$ and $\sigma2$ were calculated as 0.4 and -0.9 respectively to get the probability of failure close to 1 as $\dot{a}_{z}$ gets close to $\pm0.5$.

\subsection{RoC of Localisation Error ($\dot{\delta}_{loc}$)}

Robots sometimes encounter situations where its wheels are free to rotate, but the robot itself is stuck. Uneven terrain is an example of one such performance degrading factor. As the wheels continue spinning, the robots raw odometry estimate($x_1$) continues to change. However, other position estimates from visual odometry \cite{nister2006visual} or EKF sensor fusion ($x_2$) remain relatively constant. Such situations result in localisation errors ($\delta_{loc} = x_1 - x_2$), i.e., the difference between redundant position estimates \cite{mendoza2012mobile} of a robot. Different SLAM algorithms are robust to different levels of localisation errors ($\delta_{loc}$). However, the performance of a robot deteriorates after prolonged periods of high $\delta_{loc}$. Hence, $\dot{\delta}_{loc}$ can be used as an indicator of when a robot's SLAM or localisation is compromised. While SLAM algorithms generally provide confidence measures, we use $\dot{\delta}_{loc}$ as a vital to reduce assumptions made about the robot's localisation algorithm.

During periods of low performance degradation, the localisation error is close to 0, barring small fluctuations. During periods of high performance degradation, the localisation error steadily increases. To detect such situations, we count the number of times steps $t_{event} = t$ that $\mid \dot{\delta}_{loc} \mid$ continuously takes a non-zero value. In preliminary experiments, we observed that robot failure became more likely when $t_{event}$ was between 4-5 seconds. This is heuristically encoded as a function where the probability of suffering linearly increases (with scaling constant $k=0.2$)  with the value of $t_{event}$:

\begin{align}
    P(\textit{suffering}| \dot{\delta}_{loc}, t_{event} = t) = \begin{cases}
    k\cdot t  & \text{if } t \in [0, 5],\\
    1  & \text{if } t \geq 5.
    \end{cases}
    \label{eq: logistic}
\end{align}

\subsection{Robot Velocity ($\dot{x}$)}

A robot's velocity is a salient indicator of performance degradation. During periods of low performance degradation, a robot velocity is constant unless acceleration or deceleration is required to turn, change directions, or around way points. This constant value is generally pre-set by the manufacturer or set by the operator before use. Navigation errors, SLAM algorithm limitations and hardware issues commonly cause a robot to halt during task execution, thereby causing a sharp drop in velocity. Alternatively, motor malfunctions and braking issues cause a robot to accelerate for long periods, thereby exceeding its standard operating velocity. 

For this vital, the robot velocity is calculated by differentiating successive EKF fused position estimates of the robot. The probability of suffering is calculated as a function of the number of seconds ($t_{event} = t$) where a robots velocity is continuously trivial (i.e., close to 0), or exceeds the robot's max speed ( 1.0 m/s for a Clearpath Husky). That is, we count the number of seconds where $\dot{x} \leq \mid 0.01 \mid \text{or} \ \dot{x} \geq \mid1.0\mid$. Our experiments have shown that as the value of $t_{event}$ is generally below 3-4 seconds when the robot is facing low performance degradation. Accordingly, we encode the probability of suffering as the following sigmoid function with $a = 1.5$ and $b=2.5$ such that the probability of suffering increases when $t_{event}$ is higher than 3 seconds.

\begin{align}
    P(\textit{suffering}| \dot{x}, t_{event} = t) = \frac{1}{1+\exp\left(\left(-a \cdot t_{event}+a\cdot b\right)\right)}
    \label{eq: sigmoid_d_2}
\end{align}

\subsection{Laser Scanner Noise Variance($\sigma^2_{noise}$)}

This vital detects situations where laser scanner noise impairs a robot's ability to perceive, map, or navigate its surroundings. Noisy readings create inaccurate representations of a robot's surroundings, thereby increasing the likelihood of collisions, sub-optimal path planning, and robot failure. The methods for evaluating noise estimation or the robustness of SLAM algorithms to different types and levels of laser noise are beyond the scope of this study. Instead, we focus on the effect of additive white Gaussian noise on a Husky robot that uses the ROS navigation stack\cite{tbd}. The laser scanner measurement array is first rearranged as a square gray scale image. We use the noise variance ($\sigma^2_{noise}$) of this image as an estimate of the total laser scanner noise \cite{immerkaer1996fast}. The $\sigma^2_{noise}$ value is then calculated by convolving the image with a $3x3$ mask and applying summations on the resultant matrix. Preliminary experiments with a Husky robot showed that low noise ($\sigma^2_{noise} \approx 0.7$) had little to no effect on its navigation, and as the noise increased to $\sigma^2_{noise} \approx 1.4$, the robot's likelihood of halting or failing increased. This effect of noise variance values between 0.5 to 1.5 on the robot is captured by designing $P(\textit{suffering}|\sigma^2_{noise})$ as a sigmoid function (similar to equations \ref{eq: sigmoid_d} and \ref{eq: sigmoid_d_2}) defined over $\sigma^2_{noise}$, with constants $a=5$ and $b=1$.

\subsection{Applying the Robot Vitals Framework to Different Cases} %Addition

% To apply this framework to different use cases, the question that needs to be answered is \say{What are the sensor readings when robot's performance starts to degrade, and how to detect them?}. The answer is generally obtained using empirical data from simulations, real-world experiments, or even from a robot's/sensor's specification sheet. Additionally, the exact nature of task or the environment do not affect the vitals, as long as a) the task is within a navigational context, and b) the effect of the environment on robot performance degradation is captured by at least one of the vitals. Consider the case of applying this framework to a Turtlebot with a laser range finder carrying out an exploration task in a noisy environment while traversing rough terrain. The value of jerk ($\dot{a}_z$) required to topple the Turtlebot may be much less than the Husky's (comparatively sturdier). Similarly, if the SLAM algorithm used is more robust, the minimum value of ($\sigma^2_{noise}$) required to degrade performance can be determined in simulation by varying the levels of noise. Therefore, once events of interest are detected in each vital, the shape of the $P(\textit{suffering})$ can be adjusted based on the robustness of the robot to performance degradation, by changing the parameters.

To apply this framework to different use cases, the question \say{What are the sensor readings when robot's performance starts to degrade, and how to detect them?} needs to be answered. This is answered by observing empirical data from simulations, real-world experiments, or even from a robot's/sensor's specification sheet. Factors like the nature of task and environment of operation do not affect the application of this framework as long as a) the task is within a mobile robot navigational context, and b) the effect of the environment on robot performance degradation is captured by at least one of the vitals. 

Consider the case of applying Robot Vitals to a Turtlebot. After pre-processing robot data to create the vitals, the thresholds for event detection need to be adjusted. For example, the jerk ($\dot{a}_z$) required to topple a Turtlebot may be lesser than that of a Husky which is comparatively sturdier. Similarly, if the robot is operating in a noisy environment, the SLAM algorithm used may be more robust. In that scenario, the minimum value of ($\sigma^2_{noise}$) required to degrade performance can be determined in simulation by varying the levels of noise. Finally, once the event thresholds are fixed, the $P(\textit{suffering})$ function's shape can be modified by adjusting its constants.

\section{Robot Health}

The robot health is a scalar estimate of a robot's ability to carry out its tasks optimally without its capabilities being impaired by any performance degrading factors (i.e.,  software, hardware or environmental factors). It is important to note that `performance' in the context of this paper refers to the ability of a robot to carry out its task. A robot's health can be monitored to estimate the effect of different problematic situations on it. Performance degradation causes the health to dip, making the robot more likely to fail or perform sub-optimally. Therefore, a robot's health can help determine if operator intervention or corrective actions are required to improve a robot's performance or prevent it from failing.

Robot health combines the effect of several performance degrading factors into a single meta-metric. This is analogous to expert ensembles \cite{dietterich2002ensemble}, or the opinion of multiple experts on different aspects of a problematic situation. Exploring the merits and demerits of different methods of combining these expert opinions (e.g., sum, product, context aware weighted averages) is beyond the scope of this paper. Here, the effect of different aspects of performance degradation are quantified using the probability of suffering given the robot vitals (please refer section \ref{sec:RV}). The total probability of robot suffering at time $t$ is calculated as follows:

\begin{align}
    P(\textit{suffering})\big|_t = \eta \sum_{v \in V^{t}} P(\textit{suffering}|v)P(v)|_t
    \label{eq:Prob_Suffering}
\end{align}

Where $v$ is any robot vital from $V = \{\dot{d}_{g},\dot{a}_{z},\dot{\delta}_{loc},\dot{x},\sigma^2_{noise}\}$ at time $t$, and $\eta$ is a normalisation constant. For the sake of simplicity, the probability of observing each vital $P(v)$ is assumed to be 1, i.e., a perfect observation model. In future work, this can be replaced by sophisticated models that use context-aware weightages, incorporating factors such as communication delays or component wear and tear.

% AS A SIMPLE EXAMPLE, TAKING A PRODUCT - MULTIPLYING SEVERAL METRICS, EACH OF WHICH IS EXPRESSED AS A PROBABILITY BETWEEN 0-1, GIVES VERY DIFFERENT BEHAVIOURS THAN E.G. AVERAGING THEM (WHICH IS SUMMING PLUS A RE-WEIGHTING). SO YOU NEED TO THINK ABOUT WHAT YOU WANT THE HEALTH METRIC TO DO IN RESPONSE TO DIFFERENT SITUATIONS IN WHICH DIFFERENT VITALS TAKE DIFFERENT VALUES. DO YOU WANT TO SMOOTH OVER ONE FREAK VITAL BEING NOISY, OR DO YOU WANT AN INDIVIDUAL VITAL TO BE ABLE TO SOUND THE ALARM EVEN WHEN OTHER VITALS REMAIN SILENT?

Information entropy is a standardised metric used to quantify the amount of information uncertainty or `surprise' in a random variable's possible outcomes. Low entropy is observed when a robot is operating under little to no performance degradation (i.e. 'normal' operating conditions). Sudden exposure to severe performance degrading factors, or a steady rise in the level of performance degradation faced by the robot will cause the entropy to rise. When the performance degrading factors are mitigated and the robot returns to normal operating conditions, the entropy can fall again. To intuitively associate high and low health with high and low performance degradation respectively, we use the additive inverse of the information entropy as the robot health. The robot health between two time intervals $t_1$ and $t_2$ is calculated using information entropy as follows:

\begin{align}
    H^{t_1:t_2} = \sum_{t=t_1}^{t= t_2}   P(\textit{suffering})\big|_t \ \cdot log( P(\textit{suffering})\big|_t)
    \label{eq:Health}
\end{align}

\section{Experimental Validation}

The goal of the experiments carried out in this paper was to determine if a robot's runtime performance degradation can be estimated using the robot's vitals and health. This is accomplished by introducing different levels of performance degradation in an autonomous mobile robot navigation task, and their effects on the robot's health and overall task performance. Two experiments were carried out using a Husky Robot, the first in simulation (referred to as Exp. I) and the second using the real robot (Exp. II). In each of these experiments, the robot was tasked with autonomously navigating from point A to B in the arenas as shown in Figs \ref{fig:setIArenas} and \ref{fig:setIIArenas}. The repository containing the ROS code for robot vitals and robot health, and any code necessary to replicate the experiments described, is provided under MIT license\footnote[2]{ \url{https://github.com/anikethramesh/robotVitals}}.

In both experiments, the robot uses the ROS navigation stack \cite{tbd} with the dynamic window local planner and a global planner that uses Djikstra's algorithm. Rotate recovery was disabled to minimise confounding factors that affect the experiment. The robot is not given any prior information about the map, terrain, boundary conditions or any of the performance degrading factors. If a robot aborts its navigation plan due to terrain adversities, laser scanner noise or a path planning timeout, the navigational goal is reset. However, if the robot is stuck, unable to find a path to the goal or aborts navigation for 30 seconds despite resetting the goal, the experimental trial is terminated.

Three common field repairable performance degrading factors observed in adverse environments were used in the experiments: High Friction (HF) terrain, uneven terrain, and laser noise. Laser scanner noise degrades the robot's ability to perceive its environment, thereby introducing localisation and navigation errors. Such errors would influence the values of the vitals $\dot{d}_{g}$, $\dot{x}$, and $\sigma_{noise}^2$. Traversing uneven terrain causes instability. Changes in elevation may result in the laser scanner detecting the ground as an obstacle, thereby degrading navigation. Such errors would affect the values of $\dot{d}_{g}$, $\dot{x}$, $\dot{a}_{z}$, and $\dot{\delta}_{loc}$. Finally, robots face difficulty turning and moving smoothly on HF terrain. Robots may slip, skid, or even halt on such surfaces. The values of $\dot{d}_{g}$, $\dot{x}$, and $\dot{\delta}_{loc}$ are affected in such cases. Different intensities of these performance degrading factors were combined to create multiple experimental conditions (see Table. \ref{fig:SimParam}).

\begin{table}
  \centering
  \vspace*{0.8cm}
  
 \includegraphics[width=0.4\textwidth]{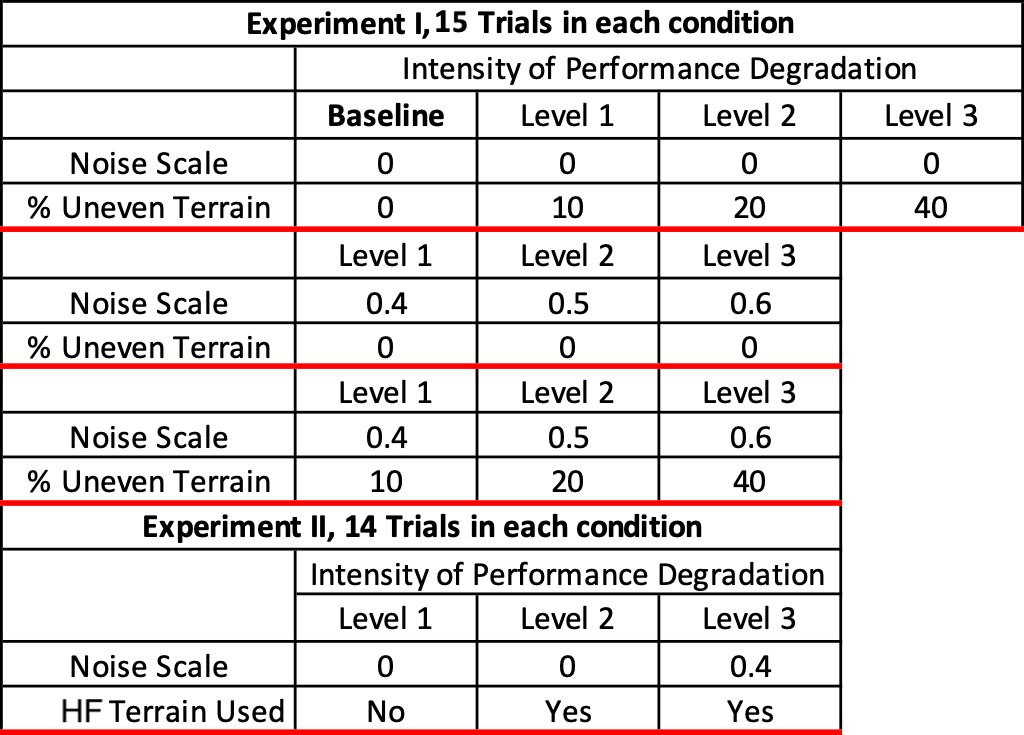}
    \caption{The different conditions tested in the experiments.}
    \label{fig:SimParam}
\end{table}

A total of 15 trials were carried out for each degradation level in Exp I and 14 trials in Exp II. After each experimental trial, we measured the time taken to complete the task $T_{comp}$. Additionally, we extracted the robot health, measured during runtime, and calculated the average robot health for each experimental trial. Performance degradation increases the time taken by a robot to complete its task. Hence, we use the time to complete the navigation task ($T_{comp}$) as an objective post-hoc measure of performance degradation. By increasing the level of performance degradation in the task, we hypothesise that: 1) The value of $T_{comp}$ increases, 2) the average robot health over the experiment runtime decreases and, 3) the average robot health and the $T_{comp}$ are inversely correlated (i.e., as the robot health decreases, the value of $T_{comp}$ increases).

\subsection{Experiment I}
\begin{figure}[h]
  \centering
%   \vspace*{0.8cm}
 \includegraphics[width=0.4\textwidth]{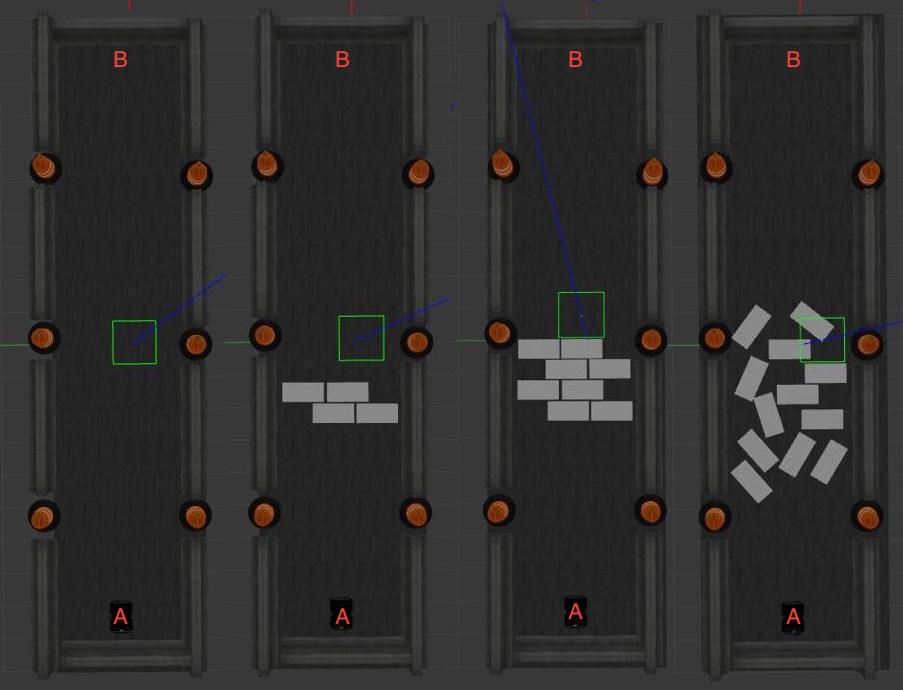}
    \caption{Gazebo Arenas used for experiments. From left to right, uneven terrain covers 0\%, 10\%, 20\%, and 40\% of the area respectively.}
    \label{fig:setIArenas}
\end{figure}

Gazebo, a high-fidelity robotics simulation with a realistic physics engine, was used for Exp I. The robot was equipped with wheel encoders for odometry, an LMS-111 LIDAR scanner and a UM6 IMU sensor. 7 seconds after the start of each trial, varying degrees of random additive Gaussian white noise were introduced into the laser scanner to degrade the robot's localisation using the Box-Muller transformation \cite{box1958note}. Laser noise was turned off after 7 seconds. The duration of noise was chosen heuristically, such that enough performance degradation was induced during runtime without causing a robot failure. To control the magnitude of the noise, the standard deviation of the Gaussian kernel is multiplied with a noise scale. The different values of the noise scale used for the experiments are listed in Fig. \ref{fig:SimParam}. For Exp I, we also designed four different arenas with uneven terrain as shown in Fig. \ref{fig:setIArenas}. These arenas had 0\%, 10\%, 20\% and 40\% of their total area covered with the `FRC 2016 Rough Terrain' gazebo block. 

\subsubsection*{Results of Experiment I}

% However, the number of performance degrading factors or their intensity 
% the performance degradation experienced by the robot does not steadily increase with the levels of performance degradation 

% There is no observable relation between the number of performance degrading factors used in a task and the resultant performance degradation observed. This indicates that performance degradation is not additive in nature.

% This indicates that quantization of performance degrading factors into different levels may not be possible 

\begin{figure}[h]
  \centering
 \includegraphics[width=0.48\textwidth]{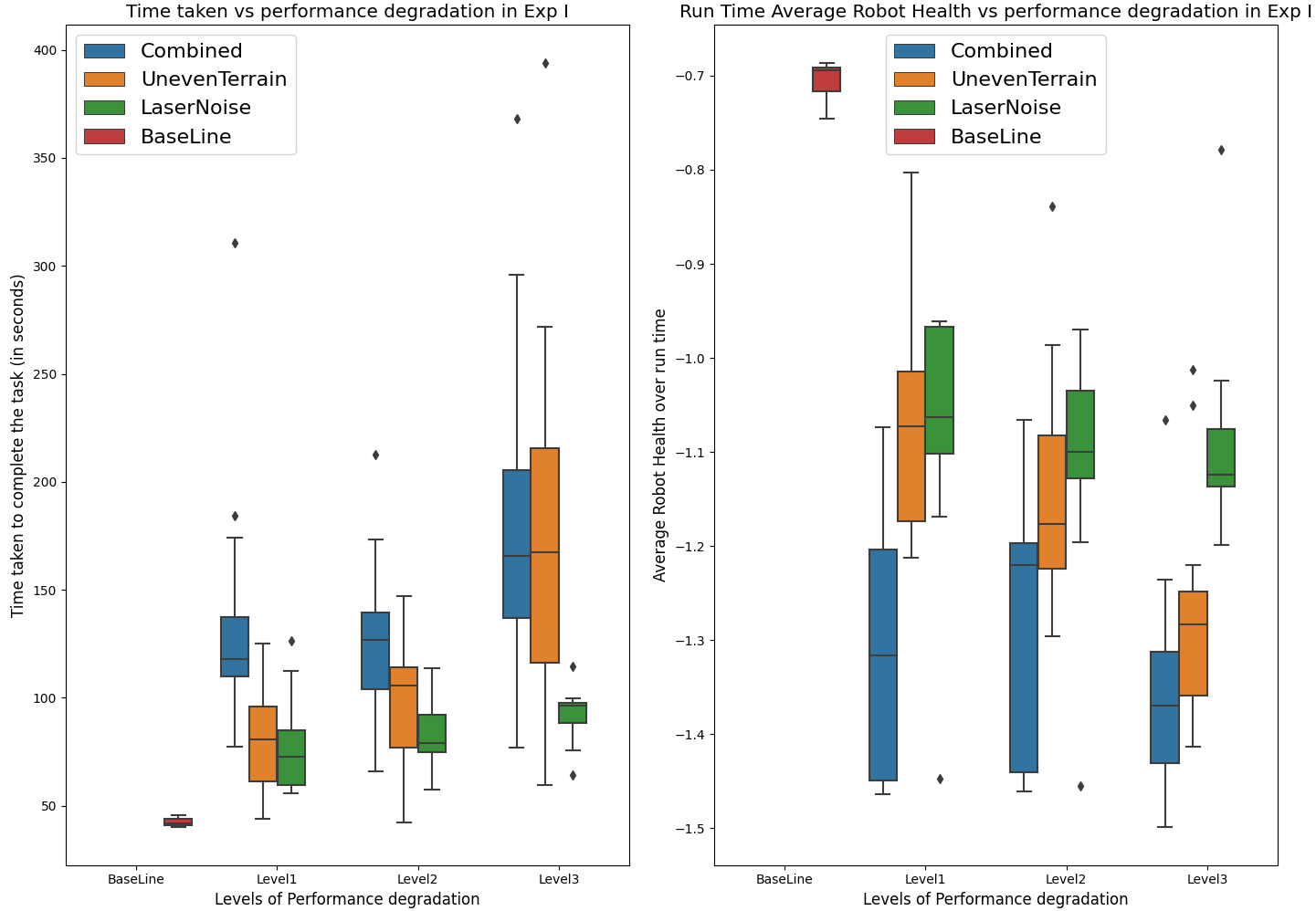}
    \caption{Exp I: \textbf{Left:} $T_{comp}$ (lower is better); \textbf{Right:} Average robot health (higher is better). The diamonds represent outliers.}
    \label{fig:setI_HTPlots}
\end{figure}

% For each trial, robot health values observed during runtime were summed and divided by the respective $T_{comp}$ (one sample per second) values to obtain the average robot health. 
The $T_{comp}$ and the average robot health values for each trial sorted by the different levels of performance degradation are plotted in Fig. \ref{fig:setI_HTPlots}. Without any performance degrading factors, the average value of $T_{comp}$ is 40 seconds and the average robot health during runtime is $-0.7$. Performance degradation increases the average time taken by the robot to complete the task and reduces the average health during runtime. However, the range of $T_{comp}$ and average robot health taken for different levels of performance degradation varies. The combination of laser noise and uneven terrain results in more performance degradation than either of the factors individually. However, experimental evidence suggests that the effect of multiple performance degrading factors on robots may not always be additive in nature, and that quantising the effects of performance degrading factors into different levels is not trivial. However, their intensities can be measured in relative terms. Among different performance degrading factors, laser noise induces the lowest level of degradation as it has caused the smallest increase in $T_{comp}$ from baseline performance. The presence of noise and uneven terrain results in the lowest range of average health values (-1.22 to -1.484). Lastly, a Spearman's rank correlation test showed a strong and significant negative correlation ($p<0.001$, $\rho= -0.93$) between $T_{comp}$ and the average robot health.
% for each trial.

\subsection{Experiment II}

\begin{figure}[h]
  \centering
  \vspace{0.3cm}
 \includegraphics[width =0.47\textwidth]{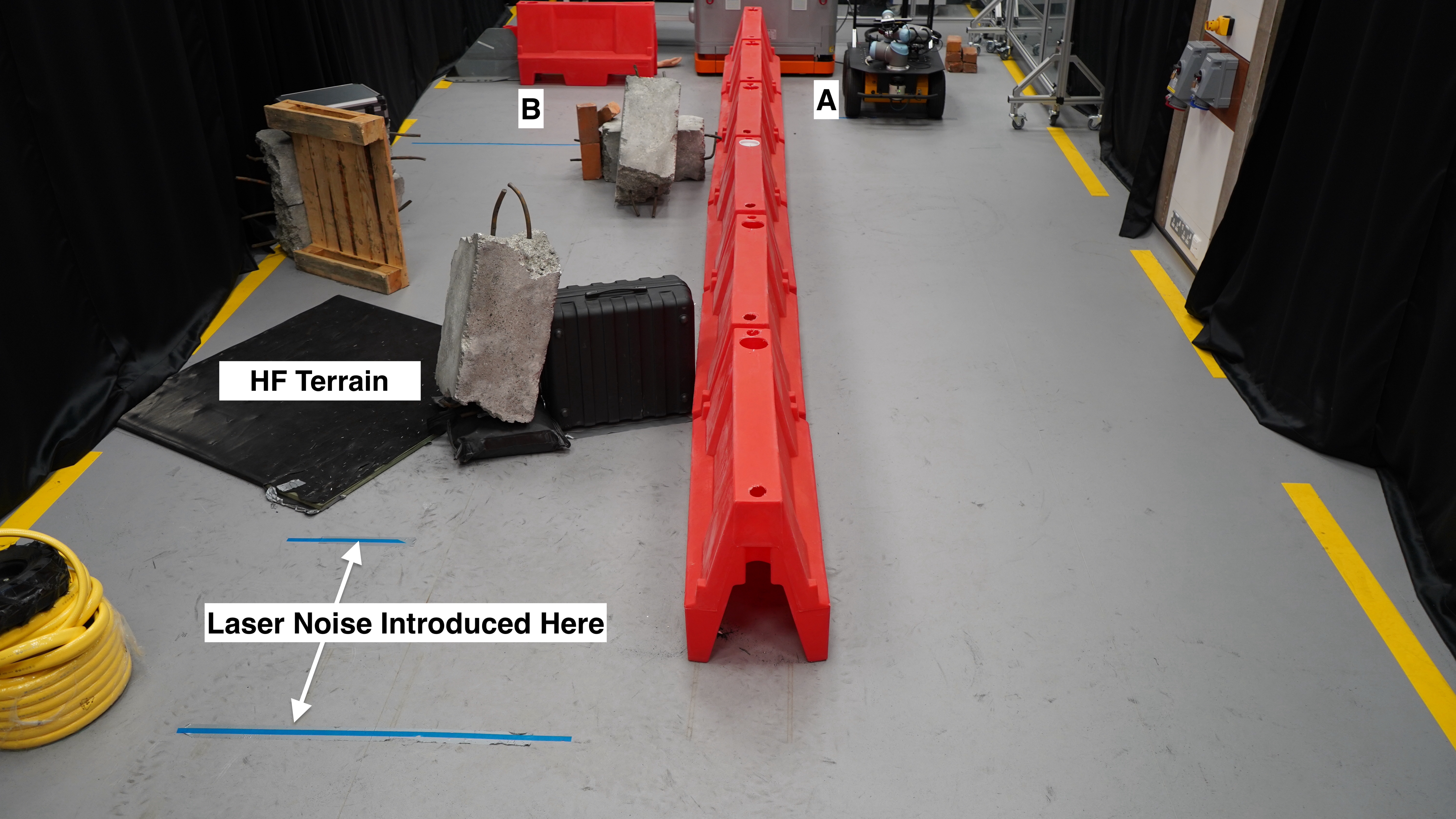}
    \caption{The arena in Exp II: Points A and B, marked with blue tape, denote the start and goal positions. The different performance degrading factors used are annotated.}
    \label{fig:setIIArenas}
\end{figure}

Exp II was carried out using a real robot in the experimental setup depicted in Fig. \ref{fig:setIIArenas}. The robot used in this experiment did not have an IMU and hence $\dot{a}_z$ values were not calculated. The robot health was calculated  using $V = \{\dot{d}_{g},\dot{\delta}_{loc},\dot{x},\sigma^2_{noise}\}$ on equation \ref{eq:Prob_Suffering}. To avoid risking hardware damage from the robot tipping over, uneven terrain was not used for the real robot experiments. Instead, HF terrain and obstacles were used to degrade the ability of the robot to turn and move smoothly. A square tile wrapped with a high friction rubber mat was used as HF terrain for this experiment. This tile was not fixed on the floor, as slippage of the tile during robot turns induces localisation errors thus adding to the performance degradation. The arena used for Exp II, and the position of the HF terrain is shown in Fig. \ref{fig:setIIArenas}. Laser noise was introduced in the annotated area for a period of 5 seconds. The locations where performance degradation was introduced were marked and kept constant.

% \begin{figure}[h]
%   \centering
%  \includegraphics[width =0.5\textwidth]{Corr_plotsCombined.png}
%     \caption{Spearmans Correlation Plots for Exp I (Left) and Exp II (Right).}
%     \label{fig:CorrPlots}
% \end{figure}

\subsubsection*{Results of Experiment II}
\begin{figure}[h]
  \centering
%   \vspace*{0.3cm}
 \includegraphics[width=0.45\textwidth]{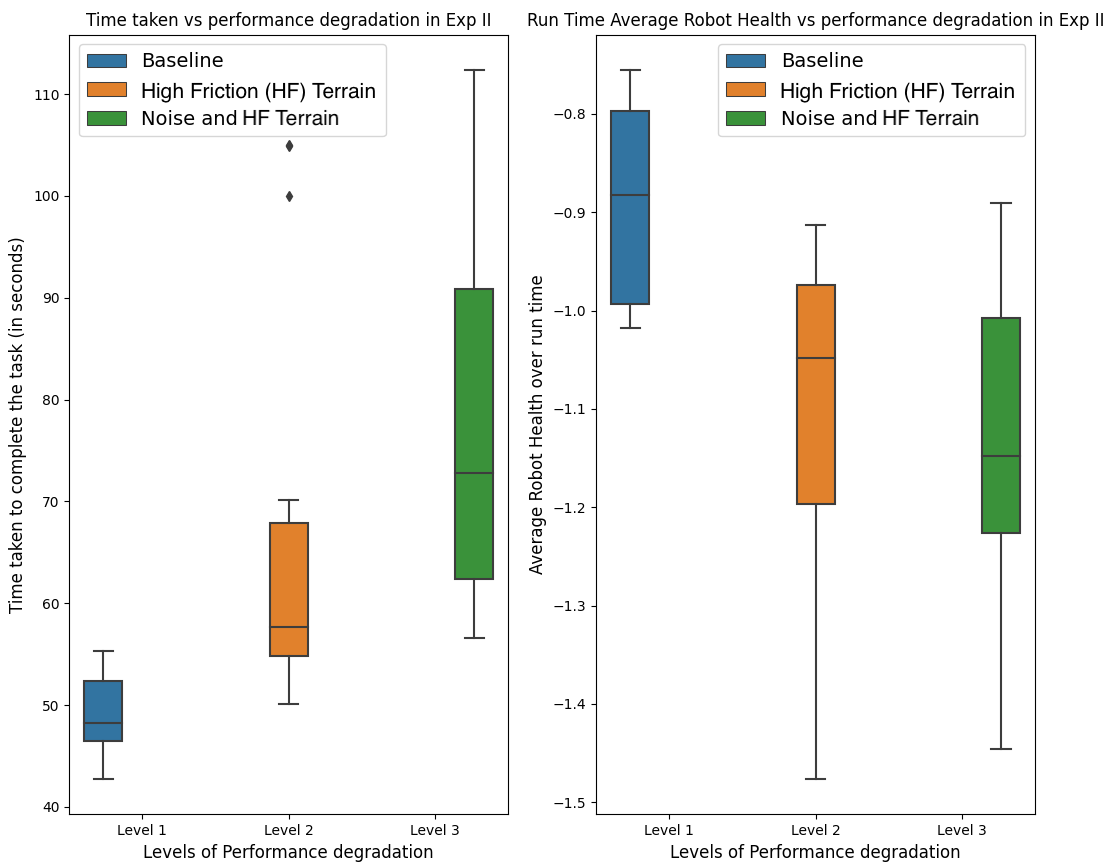}
    \caption{$T_{comp}$ (left) and average robot health (right) boxplots for Exp II. The diamonds represent outliers.}
    \label{fig:setII_HTPlots}
\end{figure}

 The average health and $T_{comp}$ values for different levels of performance degradation have been plotted in Fig. \ref{fig:setII_HTPlots}. The range of $T_{comp}$ values increases with levels of performance degradation. The middle quartile of average health observed in level 3 is lower than that of level 2, however the range of values are similar. The robot was unable to complete the task in 2 trials of level 2 and 3 trials of level 3. In these trials the robot momentarily experienced a localisation error on the HF terrain, and then could not find a collision-free path, even after resetting the goal. The Spearman's Rank Correlation test showed a significant ($p<0.001$), strong negative correlation ($\rho= -0.77$) between the average robot health and $T_{comp}$ values.

\begin{figure}[h]
  \centering
  \vspace{0.3cm}
 \includegraphics[width =0.48\textwidth]{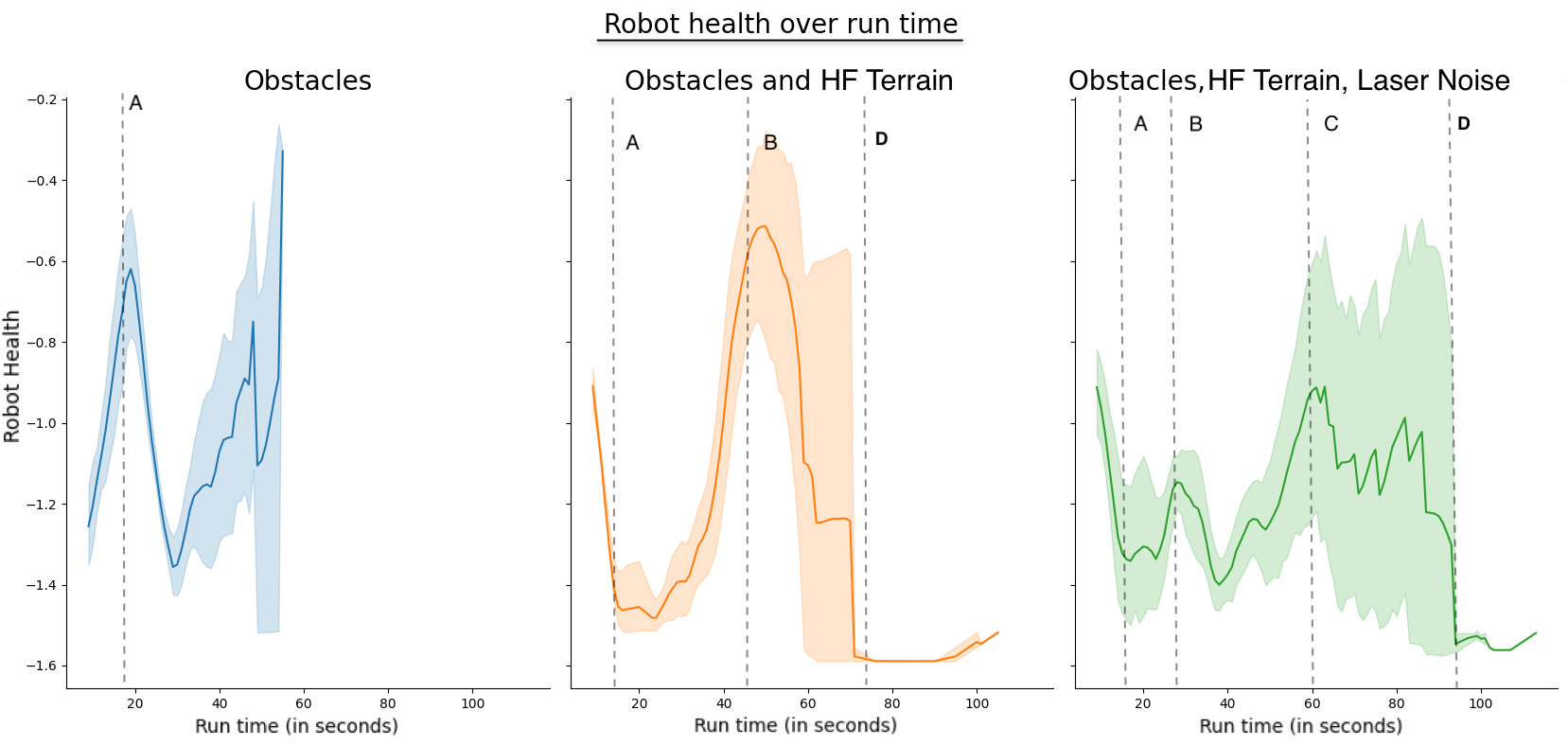}
    \caption{The health trend over runtime for all conditions in Exp II. (L to R) Level 1, Level 2, Level 3. Error bands around the lines represent the 95\% confidence interval of values. Dotted lines indicate when the robot entered the area with obstacles (A); HF terrain (B); laser noise (C);
    (D) indicates the approximate timestamp where robot failures were observed.
    }
    \label{fig:setII_HealthPlots}
\end{figure}

The instantaneous robot health for all conditions can be seen in Fig. \ref{fig:setII_HealthPlots}. Given the same experimental conditions, robotic hardware and navigation algorithms, the performance degradation induced in a robot varied in each trial. 

% The high variance in robot performance between different trials indicates that along with performance degrading factors, stochasticity in the environment of operation plays an important role in affecting the performance of the robot in any task.

As it can be seen in Fig. \ref{fig:setII_HealthPlots}, the health trend for level 1 dipped from 20 to 30 seconds' mark. In this period, the robot encountered obstacles and slowed down to create a new navigation plan. The health then continued to rise until task completion, indicating little or no performance degradation. In level 2, the health sharply dropped around the 50 second mark. This sharp drop was consistent with the robot encountering HF terrain and facing navigation errors. In level 3, the robot health trend was characterised by high fluctuations due to the effect of laser noise, obstacles and HF terrain. The introduction of multiple performance degradation factors in quick succession caused the health to stay below -0.8 through the runtime.

\section{Discussion}

% In this study, a robot navigation task was carried out under different levels of performance degradation. Three different parameters - 1)$T_{comp}$, an objective post-hoc measure of performance degradation, 2) Robot health, an online real-time measure of performance degradation and 3) The average robot health were measured for each experimental trial. 

The evidence presented show that as the level of performance degradation increases, $T_{comp}$ increases and the average robot health decreases (see Fig. \ref{fig:setI_HTPlots} and \ref{fig:setII_HTPlots}). Additionally, the robot health successfully tracks the performance degradation in real time, as seen in Fig. \ref{fig:setII_HealthPlots}. Most importantly, the strong negative correlation between $T_{comp}$ and the average robot health indicates that our proposed metric to measure performance degradation online is as effective as an objective post-hoc measure (i.e. $T_{comp}$). The provided evidence suggests that robot health can successfully estimate the instantaneous performance degradation of a robot during runtime.

An online robot performance monitoring system can be realised this framework. Such a system could detect low health (i.e. high performance degradation) situations autonomously, trigger recovery behaviours or request operator assistance. Exp II gives insights on how such a system could prove useful. In some trials of Exp II, the robot got stuck momentarily or failed when it was unable to find a collision free path through the arena. During these instances the robot's health dropped below $-1.4$. Thus, a simple threshold-based control switcher \cite{chiou2019mixed} could be used to initiate recovery behaviours. Alternatively, an operator could take control of the robot to provide it a collision free path or teleoperate it. In multi-robot applications, the robot health can also be used to prioritise robots in need of operator attention based on the severity of performance degradation. Explainable AI agents that assist operators and autonomous robots with initiating recovery behaviours, can use the vitals and health of robots to substantiate their decisions. Elaborating on the design and implementation of such control switcher is however beyond the scope of this paper.

While the health metric in Exp I used 5 vitals, Exp II did not use $\dot{a}_z$ as the robot did not have an IMU. The ability of our framework to still capture the overall effect of performance degradation on the robot demonstrated that it is generalisable, scalable and robust to the addition or removal of at least one vital. Any new indicator that qualifies as a robot's vital can be easily added to the health metric using a $P(\textit{suffering}|v)$ function that relates changes in the vital with the probability of robot's failure. This function can be derived from Monte Carlo studies, reliability studies, or expert knowledge. For example, an expert's knowledge can be used to recognise faulty sensor readings, or unusual robot behaviour in specific environments. Alternatively, a set of vitals can be found using methods like principal component analysis or machine learning. Such approaches can mine large amounts of data from robot operation to find parameters that best represent its performance degradation. Similar techniques can be used to design the $P(\textit{suffering})$ function or the health metric. However, encoding expert knowledge in this system makes it intuitively explainable in comparison to black box AIs.

The variance in some of the results was due to the stochasticity introduced by performance degrading factors in the experiments. Minor differences in LIDAR map representations, odometry, or update frequencies do not affect a robot's navigation during low performance degradation or normal operation. However, factors like laser noise distort the robot perception of its environment. In some trials this led to an increase in the likelihood of collisions and failures; in other trials the robot took sub-optimal paths towards the goal. In experiment II, small differences in the path of the robot caused it to climb on the high friction terrain at different angles. In some such cases the robot got stuck/ experienced a failure, and in others it was able to finish the task. Due to this stochasticity, an increase in the level of performance degradation in the experiments did not always induce a commensurate increase in $T_{comp}$.

\section{Conclusion}

In this paper, we proposed the framework of robot vitals and robot health to automatically detect and quantify the performance degradation experienced by mobile robots during task execution. We also outline a systematic approach to think, reason about and develop metrics that quantify a robot's ability to carry out its tasks. From systematic experiments on a simulated and real mobile robot, we conclude that the proposed robot vitals and robot health metric can effectively estimate the instantaneous run time performance degradation a robot is facing. The robot health can be used to automatically initiate recovery behaviours when a robot is likely to fail or is performing its tasks sub-optimally. Evidence show that the proposed framework is scalable and robust to the addition or removal of vitals. In future work, we aim to evaluate the utility of this framework in assisting human-initiative, robot-initiative, and mixed-initiative paradigms in variable autonomy systems \cite{Chiou2016IROS_HI,chiou2019mixed,chiou2015towards}.

% address the problem of quantifying and estimating the performance degradation in mobile robots. This framework was validated through systematic experiments both on a simulated and a real mobile robot. The experimental results demonstrate that the run-time performance degradation experienced by a robot can be estimated and represented as a single scalar value using the robot health, yielded by the combination of the robot vitals. A robot with low health is likely to fail or perform its tasks suboptimally. In future work, robot vitals and robot health show the potential to be used to monitor robots and initiate corrective actions during low health periods to improve overall task performance.

% FUTURE WORK - IN FUTURE WORK WE WILL INCORPORATE THIS VITALS-HEALTH APPROACH INTO OUR VARIABLE-AUTONOMY FRAMEWORK, CITE, CITE, CITE. WE WILL EVALUATE ITS UTILITY IN ASSISTING HI, RI AND MI CONTROL PARADIGMS.

\printbibliography

\end{document}